\documentclass{article}
\usepackage[numbers,sort&compress,longnamesfirst,sectionbib]{natbib}
\usepackage{graphicx}

\usepackage{xcolor}
\newcommand{\ignore}[1]{}
\newcommand{\remark}[1]{\textcolor{red}{(\em #1)}}
\usepackage{booktabs}

\usepackage[hidelinks]{hyperref}

\usepackage{tikz}
\usetikzlibrary{shapes.geometric}
\usetikzlibrary{positioning,shapes,shadows,arrows,calc}
\usetikzlibrary{intersections}
\pgfdeclarelayer{background}
\pgfdeclarelayer{foreground}
\pgfsetlayers{background,main,foreground}

\usepackage{pgfplots}
\usepackage{pgfplotstable}
\usepackage{colortbl}

\usepackage{algorithmic}
\usepackage{algorithm}
\usepackage[algo2e,ruled,linesnumbered]{algorithm2e}
\usepackage{mathtools}
\usepackage{amssymb}

\usepackage{xspace}
\newcommand{\TTP}{\textsc{TTP}\xspace}
\newcommand{\bitflip}{\textsc{Bitflip}\xspace}
\newcommand{\insertion}{\textsc{Insertion}\xspace}
\newcommand{\ea}{\textsc{(1+1)-EA}\xspace}

\newcommand{\PackIterative}{\textsc{PackIterative}\xspace}

\newcommand{\matls}{\textsc{MATLS}\xspace}

\usepackage{amsmath}
\DeclareMathOperator*{\mycup}{\cup}

\title{A case study of algorithm selection for the traveling thief problem
\footnote{M. Wagner was supported by the Australian Research Council (DE160100850) and by a Priority Partner Grant by the University of Adelaide, Australia. M. Lindauer and F. Hutter were supported by the DFG (German Research Foundation) under Emmy Noether grant HU 1900/2-1. M. M\i s\i r was supported by the Nanjing University of Aeronautics and Astronautics Starter Research Fund.} 
}

\author{Markus Wagner$^1$, 
Marius Lindauer$^2$, 
Mustafa M\i s\i r$^3$, \\
Samadhi Nallaperuma$^4$, 
Frank Hutter$^2$ \footnote{
$^1$ Optimisation and Logistics Group, School of Computer Science, The University of Adelaide, Australia, markus.wagner@adelaide.edu.au \newline
$^2$ Institut f\"{u}r Informatik, Albert-Ludwigs-Universit\"{a}t Freiburg, Germany, lindauer@informatik.uni-freiburg.de\newline
$^3$ Institute of Machine Learning and Computational Intelligence, Nanjing University of Aeronautics and Astronautics, China, mmisir@nuaa.edu.cn\newline
$^4$ Department of Computer Science, University of Sheffield, UK, s.nallaperuma@sheffield.ac.uk
}
}

\date{}

\begin{document}

\maketitle

\begin{abstract}

Many real-world problems are composed of several interacting components. 
In order to facilitate research on such interactions, the Traveling Thief Problem (TTP) was created in 2013 as the combination of two well-understood combinatorial optimization problems. 

With this article, we contribute in four ways. First, we create a comprehensive dataset that comprises the performance data of 21 TTP algorithms on the full original set of 9720 TTP instances. Second, we define 55 characteristics for all TPP instances that can be used to select the best algorithm on a per-instance basis. Third, we use these algorithms and features to construct the first algorithm portfolios for TTP, clearly outperforming the single best algorithm. Finally, we study which algorithms contribute most to this portfolio. 


\end{abstract}


\section{Introduction}
\label{sec:introduction}

The complexity of operations is increasing in most companies, with several interacting components having to be addressed at once. 
For example, the issue of scheduling production lines (e.g., maximizing the efficiency or minimizing the cost) has direct relationship with inventory costs, transportation costs, delivery-in-full-on-time to customers, and hence should not be considered in isolation. 
In addition, optimizing one component of the operation may negatively impact activities in other components. 

The academic traveling thief problem (\TTP) \citep{ttp2013bonyadi} is quickly gaining attention as an NP-hard combinatorial optimization problem that combines two well-known subproblems: the traveling salesperson problem (TSP) and the knapsack problem (KP). These two components have been merged in such a way that the optimal solution for each single one does not necessarily correspond to an optimal TTP solution. The motivation for the TTP is to allow the systematic investigation of interactions between two hard component problems, to gain insights that eventually help solve real-world problems more efficiently~\citep{BonyadiMNW016multicomponentOpportunities}.

Since the introduction of the TTP, many algorithms have been introduced for solving it. 
While the initial approaches were rather generic hill-climbers, researchers  incorporated more and more domain knowledge into the algorithms. For example, this resulted in deterministic, constructive heuristics, in restart strategies, and also in problem-specific hill-climbers that try to solve the TTP holistically. While the use of insights typically resulted in an increase in the objective scores, the computational complexity also increased. 
Consequently, which one of the algorithms performs best is highly dependent on the TTP instance at hand. 
To exploit this complementarity of existing algorithms, here we study the applicability of algorithm selection~\citep{rice76a} to this problem.

Specifically, after describing the TTP 
(Section~\ref{sec:ttp}) and the algorithm selection problem (Section \ref{sec:as}), we make the following contributions:
\begin{itemize}
	\item We analyze the performance of 21 TTP algorithms on the original set of 9720 instances created by \citet{Polyakovskiy2014instances} (Section~\ref{sec:benchmarks});
	\item We describe characteristics of TTP instances that can be used as ``features'' for determining the best algorithm for the instance (Section~\ref{sub:as:features});
	\item We create the first algorithm portfolios for TTP, substantially improving performance over the best single TTP algorithm (Section~\ref{sub:as:benchs}); and
	\item We analyze how complementary the algorithms in the portfolio are and which algorithms are most important for achieving good performance (Section \ref{sec:shapley_analysis}).
\end{itemize}


\section{The travelling thief problem (TTP)}
\label{sec:ttp}

The traveling thief problem (TTP)~\citep{ttp2013bonyadi} is a recent attempt to provide an abstraction of multicomponent problems with dependency among components. It combines two problems and generates a new problem with two components. In particular, it combines the traveling salesperson problem (TSP) and the knapsack problem (KP), as both problems are well known and have been studied for many years in the field of optimization. 

In this section, we motivate the TTP as an academic problem that addresses an important gap in research and then define it formally.

\subsection{Motivation}

In contemporary business enterprises the complexity of real-world problems has to be perceived as one of the greatest obstacles in achieving effectiveness. Even relatively small companies are frequently confronted with problems of very high complexity. 
Some researchers investigated features of real-world problems that served to explain difficulties that Evolutionary Algorithms (EAs) experience in solving them. For example, \citet{Weise2009whyOptDifficult} discussed premature convergence, ruggedness, causality, deceptiveness, neutrality, epistasis, and robustness, which make optimization problems hard to solve. However, it seems that these reasons are either related to the landscape of the problem (such as ruggedness and deceptiveness) or to the optimizer itself (such as premature convergence and robustness) and do not focus on the nature of the problem. \citet{michalewicz2004howToSolveIt} discussed a few different reasons behind the hardness of real-world problems, including problem size, presence of noise, multi-objectivity, and presence of constraints. Most of these features have been captured in different optimization benchmark sets, such as TSPlib~\citep{Reinelt91}, MIPlib~\citep{KochEtAl2011} and OR-library~\citep{Beasley1990}. 

Despite decades of research efforts and many articles written on Evolutionary Computation (EC) in dedicated conferences and journals, still it is not that easy to find applications of EC in the real-world. 
\citet{Michalewicz2012emperor} identified several reasons for this mismatch between academia and the real world.  
One of these reasons is that academic experiments focused on single component (single silo) benchmark problems, whereas real-world problems are often multi-component problems. 
In order to guide the community towards this increasingly important aspect of real-world optimization~\citep{BonyadiMNW016multicomponentOpportunities}, the traveling thief problem was introduced~\citep{ttp2013bonyadi} in order to illustrate the complexities that arise by having multiple interacting components.

\ignore{MW: Marius, what do you have in mind with "discussion of relevance"? Do you mean "why is the TTP intesting?"\\
ML: yes, indeed. I think the paper would be stronger if we could argue why TTP is not only an artificial problem.}

A related problem is the vehicle routing problem~(VRP, for an overview see \cite{Bell200441,Rizzoli2007}).
The VRP is concerned with finding optimal routes for a fleet of vehicles delivering or collecting items from different locations~\citep{dantzig1959,Laporte1992}. 
Over the years, a number of VRP variants have been proposed, such as variants with multiple depots or with capacity constraints. 
However, the insights gained there do not easily carry over to the academic TTP, as we consider in addition to the routing problem not only a load-dependent feature, but also the NP-hard optimisation problem of deciding which items are to be stolen by the thieves. 
For discussions on how the TTP differs from the VRP, we refer the interested reader to \cite{ttp2013bonyadi,ttp2014bonyadi}.

Despite being a challenging problem, it is often disputed whether the TTP is realistic enough because it only allows a single thief to travel across hundreds or thousands of cities to collect (steal) items. In addition, the thief is required to visit all cities, regardless of whether an item is stolen there or not. \citet{Chand2016multiplettp} discussed the shortcomings of the current formulation and presented a relaxed version of the problem which allows multiple thieves to travel across different cities with the aim of maximizing the group's collective profit. A number of fast heuristics were also proposed for solving the newly proposed multiple travelling thieves problem (MTTP). It was observed that having a small number of additional thieves could yield significant improvements of the objective scores in many cases.

\subsection{Formal Definition}

We use the definition of the \TTP by \cite{Polyakovskiy2014instances}. 
Given is a set of cities $N=\{1,\ldots,n\}$ and a set of items $M=\{1,\ldots,m\}$ distributed among the cities. For any pair of cities $i,j \in N$, we know the distance $d_{ij}$ between them. Every city $i$, except the first one, contains a set of items $M_i=\{1,\ldots,m_i\}$, \mbox{$M = \displaystyle \mycup_{i \in N} M_i$}. Each item $k$ positioned in city $i$ is characterized by its profit $p_{ik}$ and weight $w_{ik}$, thus the item $I_{ik}\sim\left(p_{ik},w_{ik}\right)$. The thief must visit all cities exactly once starting from the first city and returning back to it in the end. Any item may be selected in any city as long as the total weight of collected items does not exceed the specified capacity $W$. A renting rate $R$ is to be paid per each time unit taken to complete the tour. $\upsilon_{max}$ and $\upsilon_{min}$ denote the maximal and minimum speeds that the thief can move. The goal is to find a tour, along with a packing plan, that results in the maximal profit. 

The objective function uses a binary variable $y_{ik} \in \left\{0,1\right\}$ that is equal to one when the item $k$ is selected in the city $i$, and zero otherwise. Also, let $W_i$ denote the total weight of collected items when the thief leaves the city $i$. Then, the objective function for a tour $\Pi=\left(x_1,\ldots,x_n\right)$, $x_i \in N$ and a packing plan $P= \left(y_{21},\ldots,y_{nm_i}\right)$ has the following form:
$$Z(\Pi,P) =  \displaystyle\sum_{i=1}^n \displaystyle\sum_{k=1}^{m_i} p_{ik} y_{ik} - R \left( \frac{d_{x_n x_1}}{\upsilon_{max}-\nu W_{x_n}} + \displaystyle\sum_{i = 1}^{n-1}\frac{d_{x_i x_{i+1}}}{\upsilon_{max}-\nu W_{x_i}} \right)$$

\noindent where $\nu = \frac{\upsilon_{max}-\upsilon_{min}}{W}$ is a constant value defined by input parameters. The first term is the sum of all packed items' profits and the second term is the amount that the thief pays for the knapsack's rent (equal to the total traveling time along $\Pi$ multiplied by $R$).

Note that different values of the renting rate $R$ result in different TTP instances that might be ``harder'' or ``easier'' to solve. For example, for small values of $R$ (relative to the profits), the overall rent contributes little to the final objective score. In the extreme case $R=0$, the best solution for a given TTP instance is equivalent to the best solution of the KP component, which means that there is no need to solve the TSP component at all. Similarly, high renting rates reduce the effect of the profits, and in the extreme case the best solution of the TTP is the optimum solution for the given TSP component. 

\begin{figure}[!t]
\centering
\includegraphics[width=57mm]{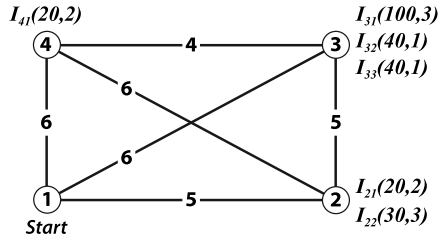}
\caption{Illustrative example for a TTP instance 
(taken from \cite{Polyakovskiy2014instances}, with permission)
}
\label{fig:Example}
\end{figure}

We provide a brief example in the following (see Figure~\ref{fig:Example}); full details are given by \citet{Polyakovskiy2014instances}. 
Each city but the first has an assigned set of items, e.g., city 2 is associated with item $I_{21}$ of profit $p_{21}=20$ and weight $w_{21}=2$, and with item $I_{22}$ of profit $p_{22}=30$ and weight $w_{22}=3$. 
Let us assume that the maximum weight $W=3$, the renting rate $R=1$ and $\upsilon_{max}$ and $\upsilon_{min}$ are set as 1 and 0.1, respectively. 
Then the optimum objective value is $Z(\Pi,P)=50$ when to tour is $\Pi=\left(1,2,4,3,1\right)$ and when items $I_{32}$ and $I_{33}$ are picked up (total profit of 80).
As the thief's knapsack has a weight of 2 on the way from city 3 back to city 1, this reduces the speed and results in an increased cost of 15. Consequently, the final objective value is $Z(\Pi,P)=80-15-15=50$.


\subsection{Algorithms for TTP}
\label{sec:algorithms}

In the following, we provide a historical overview of approaches to the TTP. 
As we shall later see, none of these algorithms dominates all others.

In the original article in which the TTP is defined, \citet{ttp2013bonyadi} used exhaustive enumeration on instances with four cities and six items in order to demonstrate the interconnected components. 
A year later, \citet{Polyakovskiy2014instances} created a set of instances with up to almost 100,000 cities and 1,000,000 items, rendering exhaustive enumeration no longer feasible.  

It were also \citet{Polyakovskiy2014instances} who proposed the first set of heuristics for solving the TTP. Their general approach was to solve the problem using two steps. The first step involved generating a good TSP tour by using the classical Chained Lin-Kernighan heuristic~\citep{chainedLK03applegate}. The second step involved keeping the tour fixed and applying a packing heuristic for improving the solution. Their first approach was a simple heuristic (SH) which constructed a solution by processing and picking items that maximized the objective value according to a given tour.
Items were picked based on a $\textit{score}$ value that was calculated for each item to estimate how good it is according to the given tour. They also proposed two iterative heuristics, namely the Random Local Search (RLS) and \ea, which probabilistically flipped a number of packing bits. After each iteration the solution was evaluated and if an improvement was noted, the changes were kept; otherwise they were ignored. 

\citet{ttp2014bonyadi} experimentally investigated the interdependency between the TSP and knapsack components of the \TTP. They proposed two heuristic approaches named Density-based Heuristic~(DH) and  CoSolver. 
DH is again a two-phased approach similar to SH from~\citet{Polyakovskiy2014instances}, and it also ignores any dependencies between the TSP and Knapsack components. 
In contrast to this, CoSolver is a method inspired by coevolution based approaches. It divides the problem into sub-problems where each sub-problem is solved by a different module of the CoSolver. The algorithm revises the solution through negotiation between its modules. The communication between the different modules and sub-problems allows for the \TTP interdependencies to be considered. 
A comparison across several benchmark problems showed the superiority of CoSolver over DH. This was especially evident for larger instances.

\citet{Mei2014} also investigated the interdependencies between the TSP and knapsack components. 
They analysed the mathematical formulation to show that the \TTP problem is not additively separable. Since the objectives of the TSP and knapsack components are not fully correlated, one cannot expect to achieve competitive results by solving each component in isolation. 
The authors used two separate approaches for solving the TTP: a cooperative coevolution based approach similar to CoSolver, and a memetic algorithm called \matls which attempts to solve the problem as a whole. The memetic algorithm, which considered the interdependencies in more depth, 
outperformed cooperative coevolution. Both works by \citet{ttp2014bonyadi} and \citet{Mei2014} highlight the importance of considering interdependencies between the \TTP components as this will allow for the generation of more competitive solutions.

\citet{Faulkner2015gecco} investigated multiple operators and did a comprehensive comparison with existing approaches. They proposed a number of operators, such as \bitflip and \PackIterative, for optimising the packing plan given a particular tour. They also proposed \insertion for iteratively optimising the tour given a particular packing. They combined these operators in a number of simple (S1--S5) and complex (C1--C6) heuristics that outperformed existing approaches. The main observation was that there does not yet seem to be a single best algorithmic paradigm for the TTP. Their individual operators, however, were quite beneficial in improving the quality of results. 
While the proposed operators seem to have certain benefits, the simple and complex heuristics did not consider the interdependencies between the TTP components, since all of these approaches were multi-step heuristics. Surprisingly, their best approach was a rather simple restart approach name S5 that combines good TSP tours with the fast \PackIterative.

\citet{Wagner2016ants} recently investigated the use of swarm intelligence approaches with the so-called Max-Min Ant System (MMAS, by \citet{Stutzle2000}). Wagner investigated the impact of two different TSP-specific local search (ls) operators and of ``boosting'' TTP solutions using TTP-specific local search. 
The resulting approaches focus less on short TSP tours, but more on good TTP tours, which can be longer. This allowed them to outperform the previous best approaches \matls and S5 on relatively small instances with up to 250 cities and 2000 items.

\citet{Yafrani2016ttp} studied and compared different approaches for solving the TTP from a metaheuristics perspective. Two heuristic algorithms were proposed, including a memetic algorithm (MA2B) and one using simulated annealing (CS2SA). The results show that the new algorithms were competitive to S5 and MATLS on a range of larger TTP instances.

Lastly, we would like to mention that no efficient complete solver for the TTP is known. One of the reasons for this appears to be the fact that even when the tour is kept fixed, packing is NP-hard~\citep{PolyakovskiyN14}. 


\section{Algorithm Selection}
\label{sec:as}

As we shall see in Section~\ref{sec:benchmarks},
no algorithm dominates all other algorithms on all instances.
One way to exploit this complementarity  of the algorithms is to use algorithm selection~\citep{rice76a,huberman-science97a} 
to select a well-performing algorithm on a per-instance base. 

\subsection{Problem Statement}
\label{sub:as:back}

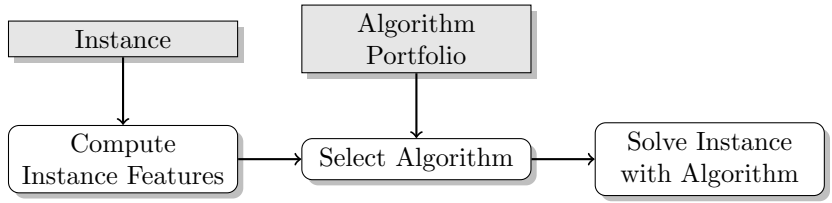
\begin{figure}[t]
\tikzstyle{activity}=[rectangle, draw=black, rounded corners, text centered, text width=8em, fill=white, drop shadow]
\tikzstyle{data}=[rectangle, draw=black, text centered, fill=black!10, text width=8em, drop shadow]
\tikzstyle{myarrow}=[->, thick]
\begin{tikzpicture}[align=center,node distance=3.9cm]
	\node (Features) [activity] {Compute Instance Features};
	\node (Instance) [data, above of=Features, yshift=-2.3cm] {Instance};
	\node (Select) [activity, right of=Features] {Select Algorithm}; 
	\node (Solve) [activity, right of=Select] {Solve Instance with Algorithm}; 
	\node (Portfolio) [data, above of=Select, yshift=-2.3cm] {Algorithm Portfolio};
	
	\draw[myarrow] (Instance) -- (Features);
	\draw[myarrow] (Features) -- (Select);
	\draw[myarrow] (Select) -- (Solve);
	\draw[myarrow] (Portfolio) -- (Select);
	
\end{tikzpicture}
\caption{Workflow of Algorithm Selection}
\label{fig:as}
\end{figure}

The algorithm selection problem is to find a mapping from problem instances to algorithms. This is realized by computing numerical characteristics -- so-called \emph{instance features} -- that describe a problem instance, and then learning a mapping from the resulting feature space to algorithms.
Figure~\ref{fig:as} shows the general workflow of algorithm selection.

We will describe instance features for the TTP later (in Section \ref{sub:as:features}), but a simple feature is, e.g., the number of cities.
Based on these instance features, we will select an algorithm from a portfolio of the 21 TTP algorithms we described in Section~\ref{sub:usedAlgorithms}
to solve the instance at hand. 

The selection step is typically realized with machine learning methods.
Based on gathered training data (i.e., instance features and performance data on training instances),
we learn a machine learning model that maps from instance features to a well-performing algorithm.

\subsection{Popular Algorithm Selection Approaches}
\label{sub:as:back}

One of the first successful algorithm selection procedures for satisfiability problems~\citep{biere-satbook}
was \textit{SATzilla}~\citep{xu-jair08a}. 
It mainly used two concepts: 
(i) Learning an empirical performance model~\citep{leyton-brown-cp02a,hutter-aij14a} to predict
the performance of an algorithm for a given instance and select the algorithm with the best predicted performance; and
(ii) Static algorithm schedules~\citep{xu-jair08a,kadioglu-cp11a,hoos-tplp14b}, which run a sequence of algorithms with a runtime budget each.
\textit{SATzilla} uses such a static schedule for ``pre-solving'', to solve easy instances without the overhead of computing features.

Other approaches include 
\begin{itemize}
	\item classification models (e.g., \textit{ME-ASP} by \cite{maratea-tplp13a}, \textit{3S} by \cite{kadioglu-cp11a}, and \textit{CSHC} by \cite{malitsky-ijcai13a}) that
directly learn a mapping from instance features to good algorithms;
	\item pairwise classification models (e.g., the more recent version of \textit{SATzilla}~\citep{xu-rcra11a}), which learns a binary classifier 
for each pair of algorithms, weighting each training instance by the performance difference between the two algorithms (and thereby emphasizing instances for which the two algorithms' performances differ a lot);
	\item unsupervised clustering (e.g., \textit{ISAC} by \cite{kadioglu-ecai10}) to partition instances in the feature space into homogeneous subsets and
then select the best-performing algorithm of the cluster a new instance is closest to; and
	\item recommender systems (e.g., \cite{MisirTechRep}) to recommend an algorithm given only partial training data.
\end{itemize}
For a thorough overview on algorithm selection procedures, we refer the interested reader to \cite{kotthoff-aim14a}. 

As was shown in the 2015 ICON challenge on algorithm selection\footnote{\url{http://challenge.icon-fet.eu/}}, 
there currently exist two state-of-the-art algorithm selection approaches.
The first is the pairwise classification version of \textit{SATzilla}~\citep{xu-rcra11a}, which won the ICON Challenge.
The second is the automatic algorithm selection method \textit{AutoFolio} system~\citep{lindauer-jair15a}.
\textit{AutoFolio} uses the flexible \textit{FlexFolio} framework~\citep{hoos-tplp14a}, which combines several different algorithm selection methods,
and searches for the best suited algorithm selection approach (and its hyperparameter settings) for a given algorithm selection scenario using 
algorithm configuration~\citep{hutter-jair09a} via the model-based configurator \textit{SMAC}~\citep{hutter-lion11a}.
For example, \textit{AutoFolio} determines whether classification or a regression approach will perform better
and in case of classification, how to set the hyperparameters of a random forest classifier~\citep{breimann-mlj01a}.
As shown by \cite{lindauer-jair15a},
\textit{AutoFolio} often chooses the pair-wise classification approach of \textit{SATzilla}, but it is more robust than other algorithm selection approaches
since it can also switch to other approaches if necessary. As a result, \textit{AutoFolio} established state-of-the-art performance on several different domains in the algorithm selection library~\citep{bischl-aij16a} and performed best on two out of three tracks of the ICON challenge.


\section{Benchmarking of TTP Algorithms}
\label{sec:benchmarks}

An important step toward the creation of algorithm portfolios is the conduct of  experiments where one determines the performance of algorithms on the available problem instances. To this end, we introduce in this section the originally defined set of TTP instances, and we outline the experimental setup and the results.

\subsection{Introduction of Benchmark Instances}

For our investigations, we use the set of TTP instances defined by~\citet{Polyakovskiy2014instances}.\footnote{As available at the TTP project page: \url{http://cs.adelaide.edu.au/~optlog/research/ttp.php}} In these instances, the two components of the problem have been balanced in such a way that the near-optimal solution of one sub-problem does not dominate over the optimal solution of another sub-problem. 

The characteristics of the original 9,720 instances vary widely. 
We outline the most important ones in the following:\footnote{For a more detailed description, we refer the interested reader to~\cite{TTPTestData,Polyakovskiy2014instances}.}
\begin{itemize}
\item The instances have 51 to 85,900 cities, based on instances from the TSPlib by~\cite{Reinelt91};
\item For each TSP instance, there are three different types of knapsack problems: \textit{uncorrelated}, \textit{uncorrelated with similar weights} and \textit{bounded strongly correlated} types, where the last type has been shown to be difficult for different types of knapsack solvers by~\cite{Martello99,Polyakovskiy2014instances};
\item For each TSP and KP combination, the number of items per city (referred to as an \emph{item factor}) is $F \in \left\{1,3,5,10\right\}$. Note that all cities of a single TTP instance have the same number of items, except for the first city (which is also the last city), where no items are available;
\item For each instance, the renting rate $R$ that links both subproblems is chosen in such a way that at least one TTP solution with an objective value of zero exists; 
\item Lastly, for each TTP configuration of the above-mentioned characteristics 
10 different instances exist where the knapsack capacity is varied.
\end{itemize}

The sheer size of this original TTP instance set makes comprehensive experimental evaluations computationally expensive and the high-dimensional space of  characteristics further complicates comparisons. 
For this reason, different researchers have selected different subsets, with each subset having (intentionally or unintentionally) a particular bias. 
For example, only the very first article by \citet{Polyakovskiy2014instances} considered the entire set of 9720 instances. \citet{yimei2014seal} focused on 30 larger instances with 11849 to 33810 cities. \citet{Faulkner2015gecco} covered a wider range using 72 instances with 195 to 85900 cities, and \citet{Wagner2016ants} used 108 instances with 51 to 1000 cities. 
Based on these individual and incomplete glimpses at algorithm performance, it is difficult to grasp the full picture. 

\subsection{Benchmark Results}
\label{sub:usedAlgorithms}

In order to establish a reliable data set for the subsequent analyses, we run existing TTP algorithms on all 9720 instances. 
This has the benefit of creating the complete picture using the same hardware and other conditions for the experiments. 

As code for most of the TTP algorithms outlined in Section~\ref{sec:algorithms} is available online, we can consider a wide range of different algorithms, which include constructive heuristics, hill-climbers, problem-agnostic and problem-specific heuristics, single-solution heuristics and cooperative coevolutionary approaches. 
In the following, we briefly list (in chronological order) the 21 considered algorithms with their original names (and, where applicable, abbreviated names in parentheses): 

\begin{itemize}
\item \cite{Polyakovskiy2014instances}: SH, RLS, EA
\item \cite{ttp2014bonyadi}: DH
\item \cite{Mei2014}: MATLS
\item \cite{Faulkner2015gecco}: S1, S2, S3, S4, S5, C1, C2, C3, C4, C5, C6
\item \cite{Yafrani2016ttp}: CS2SA
\item \cite{Wagner2016ants}: 
MMASls3 (M3), MMASls4 (M4), MMASls3boost (M3B), \\MMASls4boost (M4B).
\end{itemize}

We run all algorithms for a maximum of 10 minutes per instance.  All computations are performed on machines with Intel Xeon E5430 CPUs (2.66GHz) and Java 1.8. 

As the encountered objective scores cover several orders of magnitude, as well as positive and negative scores, we assess the quality of the algorithms using the following approach. 
For each TTP instance, we determine the best and the worst objective scores; these two values define the boundaries of the interval of observed objective scores for each instance. 
We then map actual scores from this interval linearly to $\left[ 0,1 \right]$, where the highest score is equivalent to 1. 
In case an algorithm did not produce a score for a particular instance, e.g. due to a time-out or crash, we assign to it the score of -1.

We report a performance overview across all 9720 instances in Figure~\ref{fig:performanceOnAllInstances}. At first sight, it appears that many algorithms perform comparably, since 19 of 21 algorithms achieve an average scaled performance of $>0.8$. However, this is largely because DH and SH often performed rather poorly and thus skew the scale. Figure \ref{fig:performanceOnAllInstancesNoSHDH} zooms into the remaining 19 algorithms' performance, showing that, on average, S5 and the algorithms starting with C and M perform well. 

These figures provide only a first indication, since the instance set they are based on contains many small instances (which biases this performance comparison such that algorithms performing well on small instances are favored), and some algorithms do not finish (which reduces their average scores). 
In particular, the following algorithms did not always produce solutions given the time limit: MATLS (204 unsolved instances), M3 (721), M4 (720), M3B (1342), M4B (1316), and CS2SA (6284). To the best of our knowledge, the first five of these suffer from long subroutines that keep them from stopping after the time limit is reached, while CS2SA crashes on these instances. 
We also note that the algorithms starting with M dominate on smaller instances, and that S5 performs well on larger instances. 
More detailed analyses will be presented later.

We have made the performance data set publicly available: CSV format at \url{http://cs.adelaide.edu.au/~optlog/research/ttp.php}.

\pgfplotscreateplotcyclelist{BFNWcolors}{
    red,only marks,every mark/.append style={fill=red},thick,mark size=1.7,mark=*\\
    blue!80!black,only marks,every mark/.append style={fill=white},thick,mark size=1.5,mark=square*\\
    black,only marks,thick,mark size=2.3,mark=star\\
    green!60!black,only marks,every mark/.append style={fill=green!60!black},thick,mark=triangle*\\
    violet,only marks,every mark/.append style={fill=violet},thick,mark=diamond*\\
	orange,only marks,every mark/.append style={fill=white},thick,mark=pentagon*\\
    gray,only marks,every mark/.append style={fill=gray},mark size=2,mark=x\\%
}

\pgfplotsset{cycle list name=BFNWcolors,
    width=110mm,height=30mm,
    tick label style={font=\scriptsize},
    label style={font=\small},
    title style={font=\small},
axis line style={opacity=0}, 
tick pos=left, 
ylabel style={align=center},
    xtick={1,2,3,4,5,6,7,8,9,10,11,12,13,14,15,16,17,18,19,20,21},
    xticklabels={SH,RLS,EA,DH,MATLS,S1,S2,S3,S4,S5,C1,C2,C3,C4,C5,C6,CS2SA,M3,M4,M3B,M4B},
    scaled y ticks=false,
    ymin=-0.1,ymax=1.1,	
    xmin=0,
    yticklabel style={rotate=90, inner sep=1mm},
    xticklabel style={rotate=90, inner sep=1mm},
	yticklabels={0,0.5,1},
	ytick={0,0.5,1},
    error bars/.cd,y dir=both,y explicit,
}%

\pgfplotstableread{
n	c1	c2
1	0.065436539	0.189658425
2	0.805316863	0.220471381
3	0.811733097	0.208074853
4	0.463616793	0.302214439
5	0.942769586	0.081839112
6	0.86284758	0.17558891
7	0.869213347	0.164271928
8	0.861144647	0.176032207
9	0.863616245	0.172799609
10	0.958651367	0.080364254
11	0.864830684	0.173337243
12	0.863585512	0.175737125
13	0.950617098	0.080510723
14	0.952247051	0.078897178
15	0.929153886	0.10194905
16	0.934383333	0.092012939
17	0.896751041	0.153179863
18	0.942451231	0.112312104
19	0.938296175	0.121991852
20	0.93694587	0.120478433
21	0.933369517	0.130918234
}\allInstances

\pgfplotstableread{
n	c1	c2
1	0.805316863	0.220471381
2	0.811733097	0.208074853
3	0.942769586	0.081839112
4	0.86284758	0.17558891
5	0.869213347	0.164271928
6	0.861144647	0.176032207
7	0.863616245	0.172799609
8	0.958651367	0.080364254
9	0.864830684	0.173337243
10	0.863585512	0.175737125
11	0.950617098	0.080510723
12	0.952247051	0.078897178
13	0.929153886	0.10194905
14	0.934383333	0.092012939
15	0.896751041	0.153179863
16	0.942451231	0.112312104
17	0.938296175	0.121991852
18	0.93694587	0.120478433
19	0.933369517	0.130918234
}\allInstancesNoSHDH

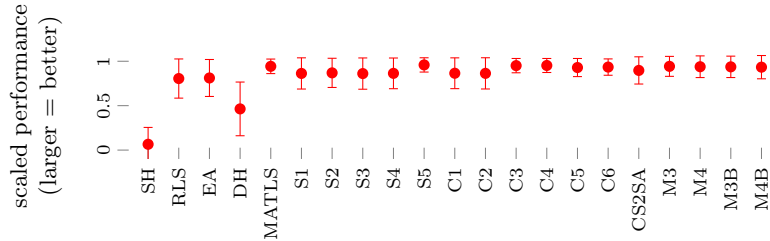
\begin{figure}[t]\centering
\begin{tikzpicture}
\begin{axis}[ylabel={\hspace{0mm}scaled performance\\\hspace{0mm}(larger = better)\vspace{-3mm}}
]
\addplot+ table[x=n,y=c1,y error=c2]\allInstances;
\end{axis}
\end{tikzpicture}%
\caption{Scaled performance of all 21 algorithms on all 9720 instances. 
}
\label{fig:performanceOnAllInstances}
\end{figure}

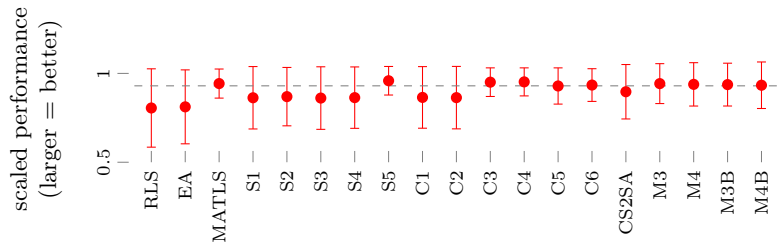
\begin{figure}[t]\centering
\pgfplotsset{
xtick={1,2,3,4,5,6,7,8,9,10,11,12,13,14,15,16,17,18,19},
xticklabels={RLS,EA,MATLS,S1,S2,S3,S4,S5,C1,C2,C3,C4,C5,C6,CS2SA,M3,M4,M3B,M4B},
ymin=0.5,ymax=1.1,	
}%
\begin{tikzpicture}
\begin{axis}[ylabel={\hspace{0mm}scaled performance\\\hspace{0mm}(larger = better)\vspace{-3mm}}
]
\addplot+ table[x=n,y=c1,y error=c2]\allInstancesNoSHDH;
\draw [dashed,gray] (5,430) -- (195,430);
\end{axis}
\end{tikzpicture}%
\caption{Scaled performance of all algorithms (except SH and DH) on all 9720 instances. Algorithms with average scores $>0.93$ (dashed line) are S5, C3, C4, C6, MATLS, M3, M4, M3B, and M4B.
}
\label{fig:performanceOnAllInstancesNoSHDH}
\end{figure}


\section{Instance Features for the TTP}
\label{sub:as:features}

For our approach to algorithm portfolio generation, we need in addition to algorithm performance data (see previous section) also data that describes problem instances. 
In total we consider 55 TTP instance features.
Of these, 47 are TSP features from previous studies on TSP~\citep{Mersmann2012tsp,Mersmann2013,Nallaperuma2013gecco, Nallaperuma2013Foga,Nallaperuma2014acoppsn}. These fall into seven groups, which we outline in the following, with the number of features in parentheses.

\paragraph{Distance Features (11).} These are based on summary statistics of the edge cost distribution. Here, we consider the lowest, highest, mean and median edge costs, as well as the proportion of edges with distances shorter than the mean distance, the fraction of distinct distances (i.e. different distance levels), and the standard deviation of the distance matrix. Also, we consider the mode frequency, quantity and mean. Finally, we used the expected tour length for a random tour, given by the sum of all edge costs multiplied by 2/(N-1).

\paragraph{Mode Features (1).} As an additional feature characterizing the distribution of edge costs, we also include its number of modes as a feature.

\paragraph{Cluster Features (6).} GDBSCAN is used for clustering where reachability distances of 0.01, 0.05 and 0.1 are chosen. Derived features are the number of clusters and the mean distances to the cluster centroids for each clusterization. 

\paragraph{Nearest Neighbor Distance Features (6).} Uniformity of an instance is reflected by the minimum, maximum, mean, median, standard deviation and the coefficient of variation of the normalized nearest-neighbor distances (nnd) of each node. 

\paragraph{Centroid Features (5).} The $x$- and $y$-coordinates of the instance centroid together with the minimum, mean and maximum distance of the nodes from the centroid.

\paragraph{MST Features (11).} Statistics which characterize the depth and the distances of the minimum spanning tree (MST). The minimum, mean, median, maximum and the standard deviation of the depth and distance values of the MST as well as the sum of the distances on the MST (which we normalize by diving it by the sum of all pairwise distances).

\paragraph{Angle Features (5).} This feature group comprises statistics of the distribution of angles between a node and its two nearest neighbor nodes: the minimum, mean, median, maximum and standard deviation.

\paragraph{Convex Hull Features (2).} The area of the convex hull of the instance reflects the ``spread'' of the instance in the plane. Additionally, we compute the fraction of nodes which define the convex hull.

\vspace{2mm}
In addition to these known TSP-specific features, we considered the following eight  features. 
The first four are features of the knapsack component and they include the capacity of the knapsack, the knapsack type, the total number of items, and the number of items per city. 
Next, we consider the number of cities as a feature. 
Lastly, as TTP-specific features we have the renting ratio, the minimum travel speed, and the maximum travel speed. 
It is important to note that these last eight do not require any processing, as they are part of the definition of the instances. 

Future investigations should include additional TTP-specific features. 
A first step towards this has been taken by \citet{PolyakovskiyN14} with their concept of ``profitable/unprofitable'' items for the special case when tours are fixed. Since we are not considering this restriction, their concept does not easily carry over. 

\ignore{THE FOLLOWING WAS AN OLD PLAN
{\footnotesize
\remark{UNTIL HERE: only description of the data... Can we include some very light analysis (plots/distributions/clustering/...)? \\ HOWEVER, the advantage of NOT having any analyses here (e.g. MM's latent feature analysis): Section 5 stays "pure" w.r.t. algorithm selection, and we would not mix the different stories here. Also: ML's approach does not use latent features (right?), so mentioning them here would not be helpful.\\
/Dropbox/Results/AlgorithmSelection/*InstanceAnalysis*\\
/Dropbox/Results/AlgorithmSelection/*FeatureImportance*\\
computation time of features in Data/tsp\_feature\_calculation\_times*, however, since we are not concerned with quick algorithm selection, there is no benefit in reporting the times}}
}


\section{Experimental Study of Algorithm Selection on TTP}
\label{sub:as:benchs}

We follow the approach of \cite{hoos-tplp14a} by studying the performance of different, well-known algorithm selection approaches.
In detail, we ran \textit{FlexFolio}\footnote{\url{http://www.ml4aad.org/flexfolio/}} (using Python $2.7.6$ and sklearn $0.14.1$)
with various approaches which \emph{simulate} the behavior of existing systems: \textit{SATzilla'09} (regression approach), 
\textit{SATzilla'11} (cost-sensitive pairwise-classification),
\textit{ISAC} (clustering) and
\textit{3S} (direct classification with $k=32$ nearest neighbors).
Since we do not optimize the runtime of our TTP algorithms but an objective score,
we cannot directly apply (pre-solving) algorithm schedules for TTP 
and hence, we focus on the classical algorithm selection approach
by selecting one algorithm per instance. 

To this end, we created an algorithm selection benchmark scenario 
in the format of the algorithm selection library (ASlib;~\cite{bischl-aij16a}) from our TTP benchmark data.\footnote{Our \texttt{TTP-2016} ASlib scenario is in the ``not\_verified'' branch of \url{http://www.aslib.net}.}
It includes the performance values for all our algorithms
and the instance features for each instance.
Furthermore, our ASlib scenario also provides the splits for a $10$-fold cross validation to obtain an unbiased performance estimate (i.e., the instances are randomly split into $10$ equally sized sets
and in each iteration, one of the splits is used as a test set to validate our algorithm selection procedure
and all others are used to train the algorithm selector; 
the overall performance of an algorithm selection procedure is then the average performance across all iterations).
With all this information saved, our ASlib scenario allows for hardware-independent reproducibility of our experiments.

\begin{table}
\centering
\begin{tabular}{l p{4cm} r}
\toprule
Simulated System & Approach & Performance \\
\midrule
\textit{Single Best} (S5) & Baseline & $0.959$\\
\textit{Oracle} & Theoretical Optimum & $1.0$\\
\midrule
\textit{SATzilla'09}-like & Regression (Lasso-Regression) & $0.966$\\
\textit{SATzilla'11}-like & Pairwise Classification (RF) & $0.993$\\
\textit{ISAC}-like & Clustering ($k$-means) & $0.989$\\
\textit{3S}-like & Classification ($k$-NN) & $0.992$\\
\midrule
\end{tabular}
\caption{Comparing different algorithm selection approaches on TTP}
\label{tab:as:benchs}
\end{table}

Table~\ref{tab:as:benchs} shows the performance of the different approaches on TTP.
Our baseline is the performance of the \emph{single best} algorithm,
i.e., always using the algorithm that performs best across all instances.
The \emph{single best} algorithm with a performance of $0.958$ is S5 (as previously shown in Section~\ref{sec:benchmarks}).
Due to the scaling of the objective scores,
the best possible score on each instance is $1$.
Therefore, the theoretical optimal performance of a perfect algorithm selection procedure (so-called \textit{oracle} or virtual-best solver) is also $1$ here,
i.e., the algorithm selector would always select best performing algorithm for each instances. 

The best-performing algorithm selection approaches are the ones of \textit{SATzilla'11}
and \textit{3S} with a nearly optimal performance of above $0.99$.
This closes the performance gap between the single best solver and the oracle 
by almost $90\%$.
\textit{SATzilla'09} and \textit{ISAC} also outperformed the \textit{single best}.
One possible reason for the good performance of algorithm selection for this application is the large instance set.
Most other instance sets studied in algorithm selection only consist of hundreds or a few thousand instances (cf. ASlib by \cite{bischl-aij16a}).
The resulting availability of more data makes the machine learning problem easier.
 
We also ran the fully automated \textit{AutoFolio} approach for a day of wallclock time on $4$ cores 
to automatically determine a good portfolio.
Since the best \textit{FlexFolio} approach (i.e., \textit{SATzilla'11}) already performed well,
\textit{AutoFolio} was only able to improve performance further by a very small margin in the $4$th decimal.
In fact, \textit{AutoFolio} also decided for the \textit{SATzilla'11} approach and
only changed the hyperparameters of the underlying random forest slightly.

\section{Analysis of Algorithm Complementarity with Shapley Values}\label{sec:shapley_analysis}

A necessary requirement for algorithm selection to perform well is 
that the portfolio of algorithms is complementary on the used instances.
A first indicator for the complementarity of the portfolio is the difference
between the \textit{single best} algorithm and the \textit{oralce} performance (see Section~\ref{sub:as:benchs}).
This difference of $0.041$ in our benchmarks appears to be small, but we have to remember that 19 of 21 algorithms achieved average scaled objective scores of $>0.8$. 

\begin{figure}[!t]
\centering
\includegraphics[width=85mm]{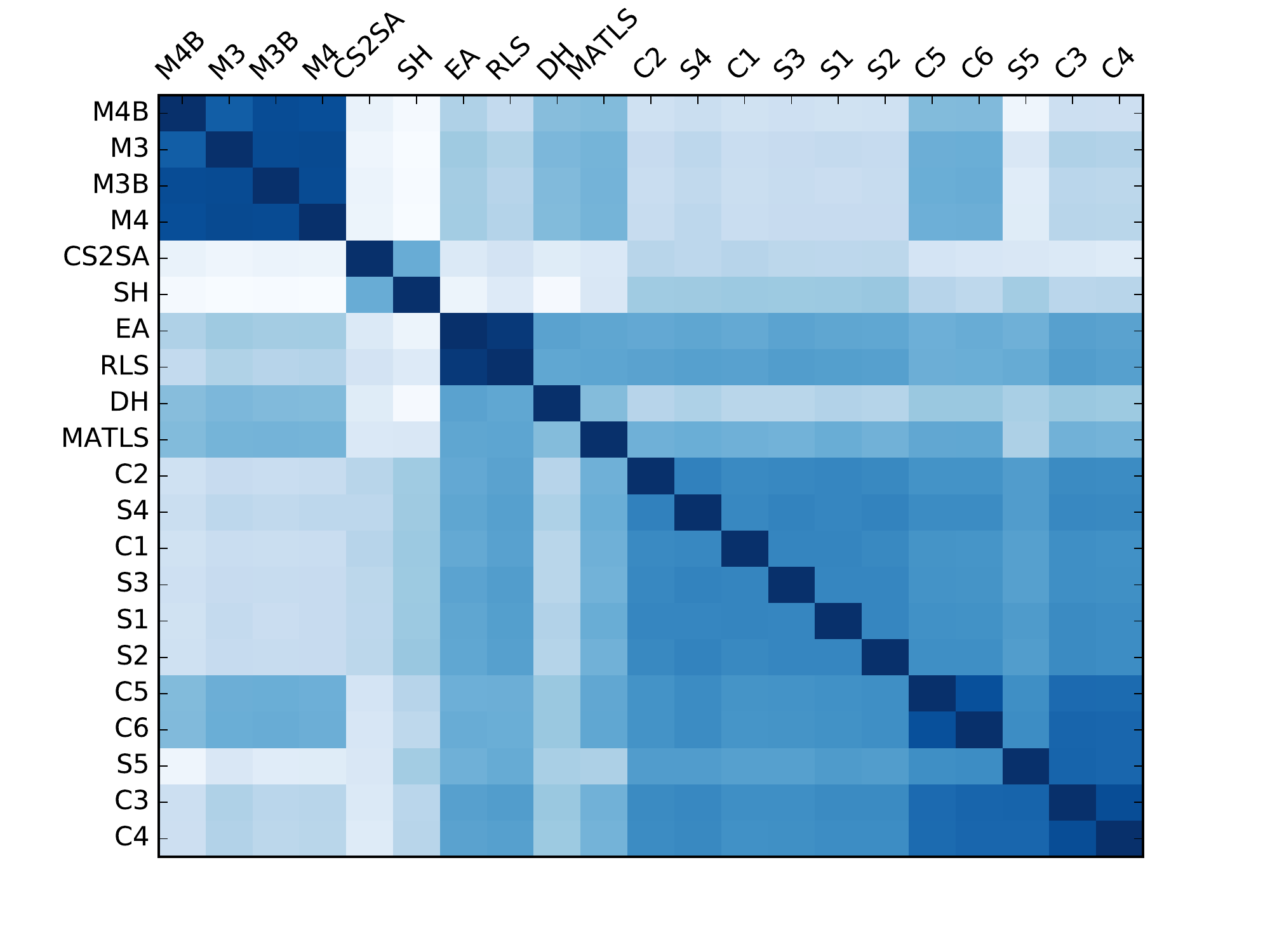}
\caption{Spearman rank coefficients in a heatmap (dark fields correspond to large correlation). The algorithms are sorted by hierarchical clustering using Ward's method.}
\label{fig:correlation}
\end{figure}

Figure~\ref{fig:correlation} shows the performance correlation across instances (based on Spearman's rank coefficients) between all pairs of algorithms.
This figure shows that the algorithms form clusters that reflect their historical development. 
For example, C* and S* fall into one cluster (all use the same fast packing heuristic), the ant-colony approaches M* form one cluster,
and early hill-climbers EA with RLS in another one. 
The algorithms CS2SA, SH, DH and MATLS are complementary to all other algorithms. 
We note that this analysis only provides insights about similarity of algorithms, 
but it is not a sufficient indicator about the applicability of algorithm selection since one of the algorithms could still dominate all other algorithms.

Another approach of assessing complementarity of algorithms is the \emph{marginal contribution} to the oracle performance~\citep{xu-sat12a},
i.e., how much the oracle performance of an existing portfolio will be improved by adding a new algorithm to it.
This approach has the disadvantage of being strongly dependent on a fixed portfolio.
To get a broader overview of an algorithm's contribution,
an extension of the marginal contribution analysis consists of using \emph{Shapley values}~\citep{frechette-aaai16a},
i.e., the marginal contribution of an algorithm to any subset of the algorithm portfolio.

\begin{figure}[!t]
\centering
\includegraphics[width=90mm]{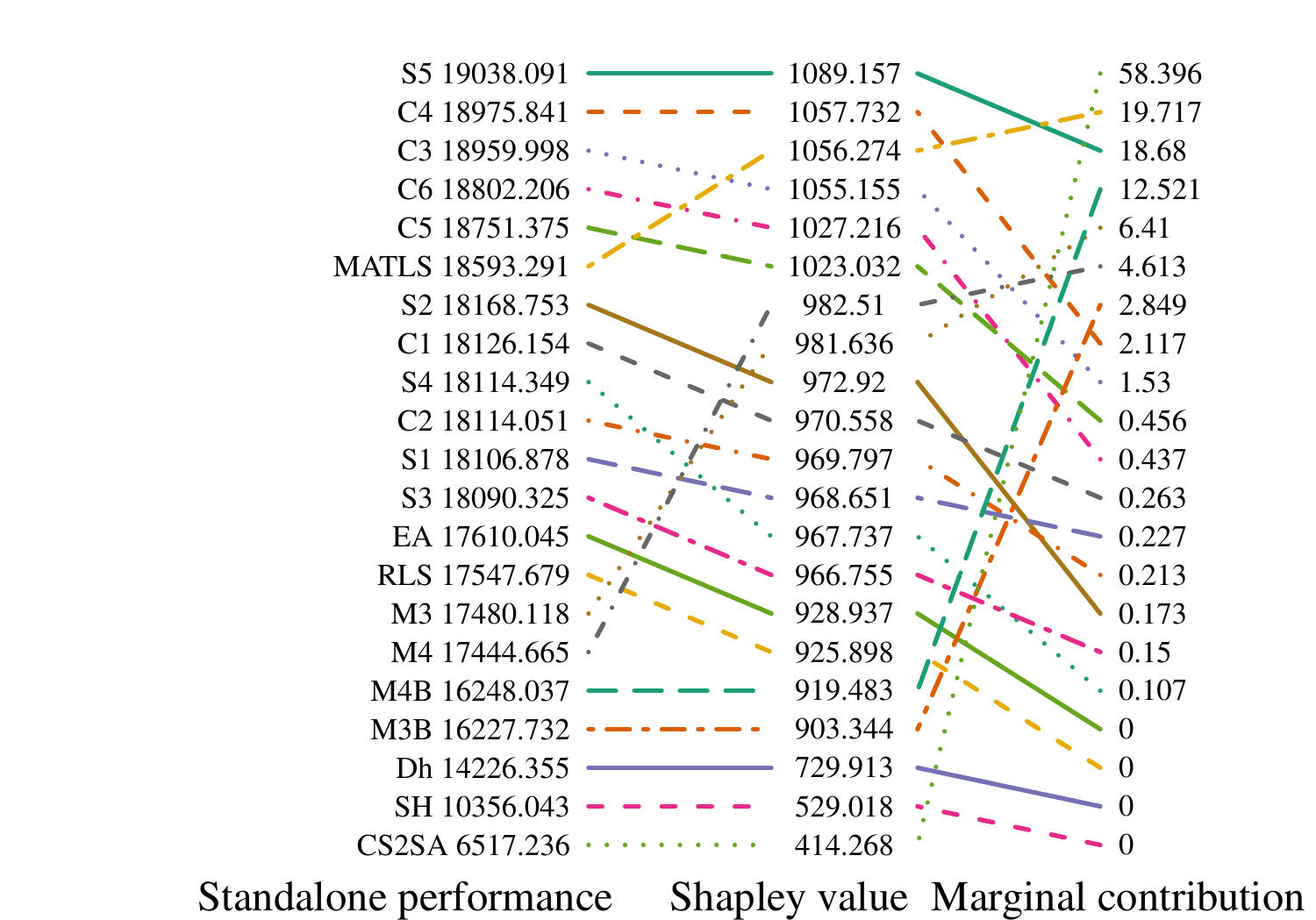}
\caption{Standalone performance, Shapley values and contributions to the oracle for all 9720 instances.}
\label{fig:shapley-21-9720}
\end{figure}

Figure~\ref{fig:shapley-21-9720} shows the ranking of the different algorithms based on their standalone performance (i.e., running only one algorithm on all instances),
the Shapley values, and the marginal contribution to the oracle performance.\footnote{The shown scores in Figure~\ref{fig:shapley-21-9720} are sums across all instances, plus an offset of ``+1'' to accommodate negative performance averages. 
}
S5 has the highest standalone performance and the highest Shapley value,
but surprisingly it is only ranked third on the marginal contribution.
Hence, S5 is a very important algorithm as a standalone and in smaller portfolios,
but does not contribute as much on top of the combination of the other algorithms as algorithm CS2SA (which has the lowest standalone performance and Shapley value
but the highest marginal contribution).
This demonstrates that CS2SA, despite its poor \emph{average} performance, performs very well on a subset of instances -- and reliably enough so for the algorithm portfolio to exploit. Once the algorithmic issues of CS2SA are fixed we expect to see significantly better average performances by this algorithm. 
The ant-colony approaches M* do not perform too well on average, but they can make useful contributions to algorithm portfolios since they perform very well on small instances.
Lastly, on the low end of the performance spectrum are two constructive heuristics DH/SH and the two uninformed hill-climbers RLS/EA. While they performed reasonably well when they were introduced, they have since then been outclassed by more informed approaches. But even the slightly informed approaches S1/S2/S3/S4, which use good TSP tours and TTP-specific packing operators, are not competitive anymore when being compared to more recent developments.

In summary, we can see that well-performing algorithm portfolios include problem-solving approaches of different complexities in order to deal with the wide range of existing TTP instances: there are swarm-intelligence approaches for small instances, memetic and multi-step heuristics for mid-size instances, and the large instances the relatively simple restart approach S5 is a good choice.


\section{Analysis of Feature Importance and Their Calculation Time}\label{sec:feature_imporance}

As the calculation of instance features forms an important step in the application of algorithm portfolios, we review the necessary calculation times in the following. In addition, we analyze which features are the most important ones for algorithm selection and we investigate how subsets of features impact computation time and portfolio performance.

To date, the established computation budget for TTP benchmarking is 10 minutes single-core CPU time per instance. For algorithm selection to be effective, and if only a single algorithm is to be run once, the calculation time of the instance features should not take up a large proportion of these 10 minutes. 
However, several of the features are computationally costly, for example, because of complete distance matrix has to be generated, or because a clustering algorithm needs to be run. 
As a consequence, the calculation time of all 55 features for a single given instance ranges from a few seconds for the smallest TTP instances to hours for the largest ones; for example, the calculations for the eil51* instances take about 2 seconds, those for the pla7397* instances are approaching 10 minutes, and the calculations for the pla33810* instances even exceed 20 hours.

To investigate which features are actually needed, as this has the potential to save significant amounts of time that then becomes available for the optimizaiton algorithm, we compute for each of the 55 features the average Gini importance~\citep{breimann-mlj01a} across all pair-wise random forests models. The results are shown in Figure~\ref{fig:features_imporance}, revealing that only a small portion of the TTP features actually matter. Interestingly, these are mostly basic knapsack features: 

\begin{figure}[!t]
\centering
\includegraphics[width=85mm]{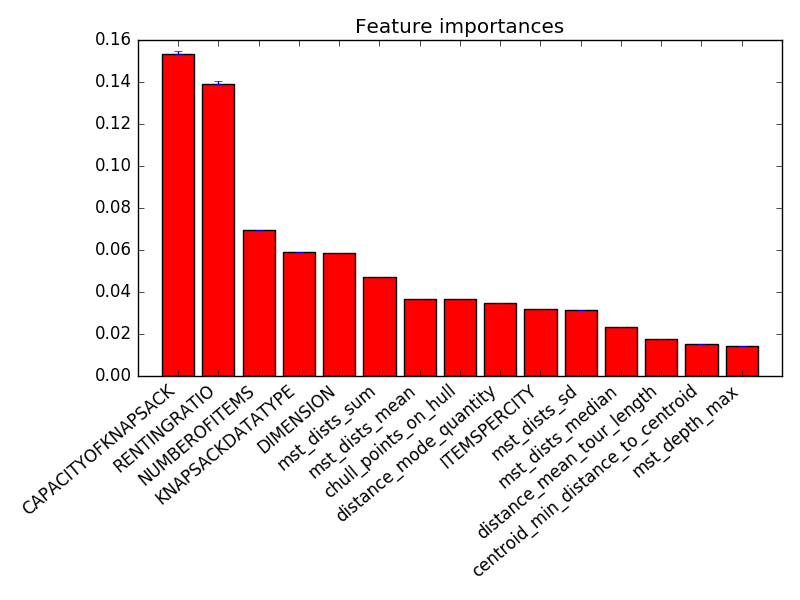}
\caption{Average feature importances of the top 15 features based on Gini importance across all pair-wise random forest models. The error-bars indicate the $25$th and $75$th percentile.}
\label{fig:features_imporance}
\end{figure}

\begin{enumerate}
\item CAPACITYOFKNAPSACK: the KP feature defining the knapsack capacity.
\item RENTINGRATIO: the TTP feature that connects the KP and the TSP.
\item NUMBEROFITEMS: the KP feature stating the total number of available items.
\item KNAPSACKDATATYPE: the KP feature stating the knapsack type.
\item DIMENSION: the TSP feature stating the total number of cities.
\end{enumerate}

As the previous portfolio investigations in Section~\ref{sub:as:benchs} were done using all 55 features, we now repeat the algorithm selection experiment using only the most important features and our best-performing approach from \textit{SATzilla'11}. The resulting performances are $0.977$, $0.980$, $0.986$, $0.988$, and $0.992$ (going from using only the most important feature to using the five most important ones). The results show that with just a small subset of the features we can achieve a portfolio performance comparable to the best one from Section~\ref{sub:as:benchs} ($0.993$).

Remarkably, all five features are given in the instance file's header, and are thus ``computable'' in constant time. 
Out of these five, CAPACITYOFKNAPSACK and RENTINGRATIO need to be defined by the instance. If NUMBEROFITEMS or DIMENSION are missing, then they can be computed by going through the instance file once and counting the total numbers of items or cities.
KNAPSACKDATATYPE is not a computable feature, as it is a parameter that was used in the generation of the instance; for our considered instance set, however, this field is always provided. Even if it is not considered, for example when using only the three most important features, we still achieve a performance of $0.986$, which is a substantial improvement over the baseline approach S5 ($0.959$).

From these experiments we see that the exploitation of immediately available instance features results in a substantial average performance increase that is comparable to a significantly more time-consuming one that requires the calculation of all 55 features.

The question now is whether we can learn even more from these outcomes. 
The visualization and interpretation of the raw outputs of the portfolios is challenging due to the large numbers of instances, features, and randomized algorithms. 
Nevertheless, let us briefly consider as an example the portfolio when only the feature CAPACITYOFKNAPSACK is used. 
Let us sort the 9720 instances according to their CAPACITYOFKNAPSACK values, and let us now consider the list of algorithms as they are selected. 
As expected, this list contains long (but not always continuous) stretches where the same algorithms are selected; in particular, the M* algortihms dominate on the tiny instances, and S5 dominates on mid-sized and large instances. 
If we do the same ordering for the algorithm selector that uses the five most important features, then the overall picture changes slightly. On the smallest $\sim$3000 instances, different complex algorithms dominate, and for the tiniest these are often the M* approaches which tend to generate the longest tours. The largest $\sim$3000 instances are typically assigned to either CS2SA (a fast implementation of search operators) or S5 (resampling solutions), which are two very different approaches.


\section{Concluding Remarks}

In this article, we presented the first study of algorithm portfolios for the TTP. We first studied the performance of 21 existing TTP algorithms on the full original set of 9720 TTP instances created by \citet{Polyakovskiy2014instances} and defined 55 instance features for TTP. Then, we studied various different approaches for the resulting algorithm selection problem, showing very substantial improvements over the single best algorithm and closing the gap between it and an omniscient oracle by 90\%. Finally, we studied which algorithms contribute most to the portfolio, finding that 
the algorithms with best average performance (e.g. the complex ones C3--C6 and MATLS, and the swarm-intelligence approaches that start with M) were quite important for the portfolio because of their performance on small and mid-sized TTP instances. Interestingly, the relatively simple heuristic S5 continues to dominate in particular on the large TTP instances and thus is one of the most important contributors to well-performing portfolios. 
Despite this general trend, the algorithm with the worst average performance, CS2SA, added substantial benefit on top of all other algorithms. 
An analysis of the feature importance revealed that the values for the five most important features can be extracted from the instance definition in constant time. The resulting portfolio that uses only this subset has a performance comparable to the one that uses all 55 features that can take hours to compute.

In future work, we aim to study 
which features make TTP instances hard for which algorithm and why, 
and whether we can identify a smaller representative subset of TTP instances to speed up future benchmarking studies.


\bibliographystyle{spbasic}      
\bibliography{aadstrings,local,aadlib,aadproc,ttp,inst}   

\end{document}